%% file: acl_latex.tex
\pdfoutput=1

\documentclass[11pt]{article}

\usepackage{acl}

\usepackage{times}
\usepackage{latexsym}
\usepackage{url}
\usepackage{xcolor,tikz}
\usepackage{amssymb}
\usepackage{booktabs,tabularx}       
\usepackage{setspace}
\newcolumntype{Y}{>{\centering\arraybackslash\hsize=0.75\hsize}X}

\usepackage[T1]{fontenc}

\usepackage[utf8]{inputenc}

\usepackage{microtype}

\input{my_commands.tex}

\title{DeepA2: A Modular Framework for Deep Argument Analysis\\ with Pretrained Neural Text2Text Language Models}

\author{
    Gregor Betz
    \\
    Karlsruhe Institute of Technology
    \\
    Karlsruhe, Germany
    \\
    \texttt{gregor.betz@kit.edu}
    \And
    Kyle Richardson
    \\
    Allen Institute for AI
    \\ 
    Seattle, WA, USA
    \\
    \texttt{kyler@allenai.org}
}

\begin{document}
\maketitle
\input{abstract}

\input{body_acl}


\section*{Acknowledgments}

We're indebted to Christian Voigt for his critical and constructive feedback throughout the DeepA2 project. %

\bibliography{bib_all}

\appendix

\input{appendix.tex}

\end{document}

%% file: my_commands.tex
\definecolor{colbrew1}{RGB}{251,180,174}
\definecolor{colbrew6}{RGB}{179,205,227}
\definecolor{colbrew3}{RGB}{204,235,197}
\definecolor{colbrew4}{RGB}{222,203,228}
\definecolor{colbrew5}{RGB}{254,217,166}
\definecolor{colbrew2}{RGB}{255,255,204}
\definecolor{colbrew7}{RGB}{229,216,189}
\definecolor{colbrew8}{RGB}{253,218,236}
\definecolor{colbrew9}{RGB}{242,242,242}

\newcommand{\maa}{\colorbox{colbrew1}{\scriptsize$\mathbf{S\!\leadsto\!{A}}$}}
\newcommand{\mab}{\colorbox{colbrew1}{\scriptsize$\mathbf{S\,R\!\leadsto\!{A}}$}}
\newcommand{\mac}{\colorbox{colbrew1}{\scriptsize$\mathbf{S\,J\!\leadsto\!{A}}$}}
\newcommand{\mad}{\colorbox{colbrew1}{\scriptsize$\mathbf{S\,R\,J\!\leadsto\!{A}}$}}
\newcommand{\mae}{\colorbox{colbrew1}{\scriptsize$\mathbf{R\,J\!\leadsto\!{A}}$}}
\newcommand{\maf}{\colorbox{colbrew1}{\scriptsize$\mathbf{P\,C\!\leadsto\!{A}}$}}

\newcommand{\mra}{\colorbox{colbrew2}{\scriptsize$\mathbf{S\!\leadsto\!{R}}$}}
\newcommand{\mrb}{\colorbox{colbrew2}{\scriptsize$\mathbf{S\,J\!\leadsto\!{R}}$}}
\newcommand{\mrc}{\colorbox{colbrew2}{\scriptsize$\mathbf{S\,A\!\leadsto\!{R}}$}}

\newcommand{\mja}{\colorbox{colbrew3}{\scriptsize$\mathbf{S\!\leadsto\!{J}}$}}
\newcommand{\mjb}{\colorbox{colbrew3}{\scriptsize$\mathbf{S\,R\!\leadsto\!{J}}$}}
\newcommand{\mjc}{\colorbox{colbrew3}{\scriptsize$\mathbf{S\,A\!\leadsto\!{J}}$}}

\newcommand{\mca}{\colorbox{colbrew4}{\scriptsize$\mathbf{A\!\leadsto\!{C}}$}}
\newcommand{\mcb}{\colorbox{colbrew4}{\scriptsize$\mathbf{O\,K\!\leadsto\!{C}}$}}

\newcommand{\mpa}{\colorbox{colbrew5}{\scriptsize$\mathbf{A\!\leadsto\!{P}}$}}
\newcommand{\mpb}{\colorbox{colbrew5}{\scriptsize$\mathbf{F\,K\!\leadsto\!{P}}$}}

\newcommand{\mfa}{\colorbox{colbrew6}{\scriptsize$\mathbf{P\!\leadsto\!{F}}$}}
\newcommand{\mfb}{\colorbox{colbrew6}{\scriptsize$\mathbf{P\,C\,O\!\leadsto\!{F}}$}}

\newcommand{\moa}{\colorbox{colbrew7}{\scriptsize$\mathbf{C\!\leadsto\!{O}}$}}
\newcommand{\mob}{\colorbox{colbrew7}{\scriptsize$\mathbf{C\,P\,F\!\leadsto\!{O}}$}}

\newcommand{\mka}{\colorbox{colbrew8}{\scriptsize$\mathbf{P\,F\!\leadsto\!{K}}$}}
\newcommand{\mkb}{\colorbox{colbrew8}{\scriptsize$\mathbf{C\,O\!\leadsto\!{K}}$}}
\newcommand{\mkc}{\colorbox{colbrew8}{\scriptsize$\mathbf{P\,F\,C\,O\!\leadsto\!{K}}$}}

\tikzstyle{vertex} = [fill,shape=coordinate,node distance=25pt]
\tikzstyle{vertex2} = [shape=circle,minimum size=14pt,color=white,node distance=50pt]
\tikzstyle{edge} = [fill,opacity=.4,fill opacity=.4,line cap=round, line join=round, line width=20pt]

\tikzstyle{roundnode} = [draw,fill=yellow,text=black,align=left,rectangle, rounded corners, line width=1pt, text width=140pt]
\tikzstyle{roundnode2} = [draw,align=left,rectangle, rounded corners, line width=1pt, minimum width=80pt]

\pgfdeclarelayer{background}
\pgfsetlayers{background,main}

\usetikzlibrary{arrows,shapes,snakes,automata,backgrounds,petri}

%% file: abstract.tex
\begin{abstract}
In this paper, we present and implement a multi-dimensional, modular framework for performing deep argument analysis (DeepA2) using current pre-trained language models (PTLMs). ArgumentAnalyst -- a T5 model \citep{raffel2020exploring} set up and trained within DeepA2 -- reconstructs argumentative texts, which advance an informal argumentation, as valid arguments: It inserts, e.g., missing premises and conclusions, formalizes inferences, and coherently links the logical reconstruction to the source text. We create a synthetic corpus for deep argument analysis, and evaluate ArgumentAnalyst on this new dataset as well as on existing data, specifically EntailmentBank \citep{dalvi2021explaining}. Our empirical findings vindicate the overall framework and highlight the advantages of a modular design, in particular its ability to emulate established heuristics (such as hermeneutic cycles), to explore the model's uncertainty, to cope with the plurality of correct solutions (underdetermination), and to exploit higher-order evidence.

[\raisebox{-1pt}{\includegraphics[height=8pt]{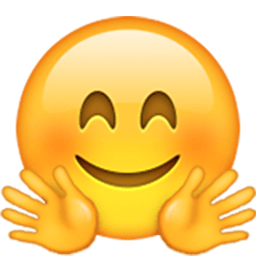}}~\href{https://huggingface.co/spaces/debatelab/deepa2-demo}{Demo}] 
[\raisebox{-1pt}{\includegraphics[height=8pt]{figs/28-hugging-face.png}}~\href{https://huggingface.co/debatelab/argument-analyst}{Model}] 
[\raisebox{-1pt}{\includegraphics[height=8pt]{figs/28-hugging-face.png}}~\href{https://huggingface.co/datasets/debatelab/aaac}{Datasets}]
\end{abstract}

%% file: body_acl.tex
\section{Introduction}

Argumentative text analysis is an interpretation method for clarifying arguments \citep{Fisher:2004cq}. Being studied in argumentation theory, logic, or epistemology, it is widely taught and applied as a key critical thinking skill in, e.g., law \citep{Alexy:1989rh}, the humanities \citep{Bruce:2011iy}, social sciences \citep{Fairclough2012}, policy advice \citep{HanssonHirschHadornRaUBook2016}, or public debate \citep{Beck_Neupane_Carroll_2019}. This paper presents a computational approach for \emph{deep argument analysis}, i.e., for \textbf{reconstructing natural-language arguments} from a given text, as in the following example \citep[adapted from][]{sep-stem-cells}:

\noindent
\begin{tabular}{@{}c@{}c@{}c@{}}
\small{\textbf{source text}}&$\leadsto$&\small{\textbf{reconstructed argument}}\\
\begin{minipage}{.48\linewidth}
\fontsize{9}{10}\selectfont
It is unethical to destroy human embryos. The most basic argument supporting this claim just stresses that it is wrong to intentionally kill innocent human beings.
\end{minipage}&&
\begin{minipage}{.47\linewidth}
\fontsize{9}{10}\selectfont
(P1) It is impermissible to kill innocent human beings.

(P2) The human embryo is an innocent human being. 

(C) \textsc{Thus}: It is impermissible to kill the human embryo.
%
%
\end{minipage}
\end{tabular}\medskip

The literature on argument reconstruction \citep[cf.][]{Feldman1998,Scholz2000,Lau:2011st,BowllKemp2014,Brun2014-BRURAF,BrunBetzRaU2016} characterizes deep argument analysis as:
\begin{itemize}
\setlength{\itemsep}{0mm}\setlength{\parskip}{0mm}
    \item a complex task involving a variety of \textbf{sub-tasks}, such as identifying reasons and conclusions in a text, formalizing sentences, checking validity of an inference, logical streamlining, or explicating implicit premises.
    \item a non-conservative, \textbf{creative task} that goes beyond mere text annotation and essentially generates a new, more transparent text. 
    \item an \textbf{iterative process} through which reconstructions are built and revised step-by-step, and the solution space is gradually explored.
    \item a hermeneutical task, guided by the \textbf{principle of charity}, which urges one to come up with an interpretation (reconstruction) as strong and plausible as possible.
    \item assuming a \textbf{normative background theory} about what constitutes a strong and plausible argument in the first place. 
    \item being affected by \textbf{severe underdetermination}, both in terms of the process and the final outcome; in particular, there typically exist rival, yet equally legitimate reconstructions of one and the same text.
\end{itemize}

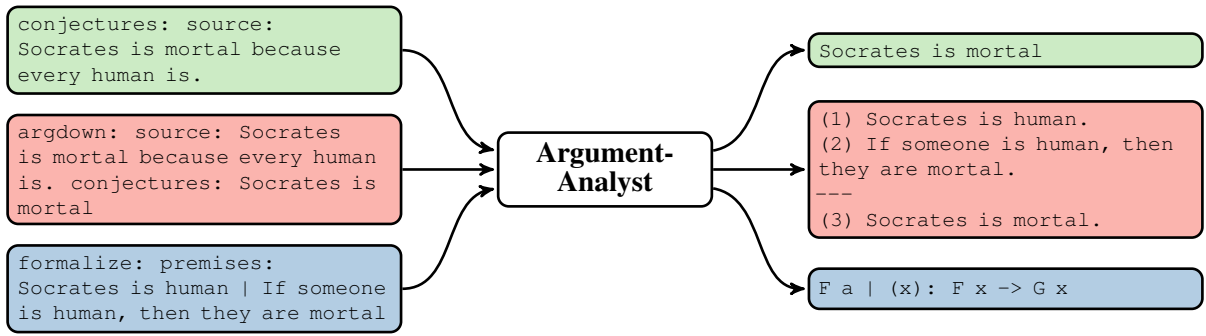
\begin{figure*}
    \begin{center}
    \input{figs/basic_design_tacl}
    \end{center}
    \caption{Example text-to-text tasks for deep argument analysis, defined by DeepA2.}
    \label{fig:basic_design}
\end{figure*}

Given these special characteristics, \emph{deep argument analysis} poses many challenges for machine models of natural language understanding. In this paper, we introduce a novel modular modeling approach for analysing complex argumentation that builds on recent pre-trained text2text transformers \cite{raffel2020exploring}. Our approach -- DeepA2 (illustrated in Figure~\ref{fig:basic_design}) -- works by systematically decomposing a complex reconstruction problem to smaller text2text sub-tasks (see Section~\ref{sec:framework}), which allows for emulating the types of interpretation strategies and heuristics studied in argument theory. Referring to the different components of a comprehensive argumentative analysis, we may also define tailor-made metrics for assessing argument reconstructions. To demonstrate the benefits of our approach, we construct a new argumentation dataset ({\small AAAC}) that exhibits several complex \emph{interpretive dimensions}, show how to map other existing datasets into our framework (Section~\ref{sec:datasets}), and train and evaluate our main model, referred to as \textbf{ArgumentAnalyst}, within DeepA2 (Section~\ref{sec:experiments}).


Our empirical results  show: 

    1. ArgumentAnalyst generates -- out-of-domain -- semantically meaningful argument reconstructions, 70\% of which are logically valid. By pooling alternative reconstructions, virtually every source text in the synthetic dataset can be reconstructed as a valid argument.

    2. Modular generation chains which emulate iterative reconstruction strategies are highly successful: they yield, in particular, a more coherent interpretation of an argumentative text, exploit the text more thoroughly, and generally outperform one-step generation as soon as problems become difficult. 
    
    3. ArgumentAnalyst outperforms \emph{EntailmentWriter} \citep{dalvi2021explaining} on difficult \emph{EntailmentBank} problems with respect to telling apart relevant premises from distractors.
    
    4. ArgumentAnalyst generates reliable higher-order evidence \citep{christensen2010higher} which can be used for diagnosing logical fallacies -- despite the fact that ArgumentAnalyst is maximally charitable and is trained to reconstruct any input whatsoever as a logically valid argument, even if the input argument, taken at face value, \emph{is} painstakingly fallacious. 

In concluding this paper, we sum-up and interpret these findings as general vindication of DeepA2's modular, multi-angular design (Section~\ref{sec:conclusion}).

\section{Related Work}



Taking \textbf{transformers as soft reasoners}, recent work, pioneered by \citet{Clark2020_TransSoftReas}, has shown that pre-trained language models (PTLMs) possess basic deductive and abductive reasoning capabilities on diverse domains \citep{banerjee2020self,betz2020critical,Bostrom2021FlexibleOF}, but are equally prone to fallacies and biases \citep{kassner2020negated,talmor2020olmpics}. Besides drawing the correct conclusion, transformers are able to generate correct reasoning chains that justify an answer, which in turn further increases answer accuracy \citep{saha2020prover,tafjord2020proofwriter,gontier2020measuring,Saha2021multiPRoverGM,dalvi2021explaining}.    

\textbf{Neural semantic parsing} uses sequence models to \emph{formalize} natural language sentences \citep{Kamath2019ASO}. \citet{Shin2021ConstrainedLM} show that PTLMs are zero-shot parsers, and that intermediate steps which rephrase and streamline the original input before parsing it to a formal language improve accuracy.

\textbf{Argument mining} is an active research field that studies computational methods for retrieving argumentative components from a text corpus \citep{Wachsmuth2017BuildingAA,Moens:2018zt,Potthast2019ArgumentSA,LawrenceReed2020}. Recently, work in this field has started to use PTLMs: \citet{EinDor2020CorpusWA} and \citet{Gretz2020ALD} succeed in retrieving relevant pro- or con-arguments for a given topic from a large corpus with a fine-tuned BERT model \citep{Devlin2019BERTPO}. Using BERT, \citet{BarHaim2020FromAT} map argumentative texts to key points that succinctly summarize the argument's gist. \citet{Akiki2020ExploringAR} explore abstractive argument retrieval by means of text generation with GPT2 \citep{Radford2019}. Similarly, \citet{Syed2021GeneratingIC} deploy BART \citep{lewis2019bart} to generate conclusions of argumentative texts on a challenging corpus compiled from Reddit and various online debate corpora. \citet{Rodrigues2020ReproductionAR}, revisiting the argument comprehension task \citep{HabernalEtAl2014,Habernal2018TheAR},  demonstrate that identifying implicit premises -- and deep argument analysis \emph{a fortiori} -- remains a hard, unsolved task. Recently,  \citet{Chakrabarty2021ImplicitPG} have shown that augmenting training data with discourse-aware commonsense knowledge improves the plausibility of automatically identified implicit premises. Such a knowledge-driven perspective is orthogonal to, and may eventually complement the logical approach adopted in this paper.




\section{Framework}
\label{sec:framework}

\subsection{Problem Definition}
\label{subsec:problem}

Deep argument analysis of a given text seeks to answer the following
\textbf{central question}: Can we make sense of the text as a presentation of a rational argument? And if so, what exactly is the argument; and how precisely is it related to the text?  

In carrying out a deep argument analysis, one explicates, rephrases and rebuilds -- even repairs -- the text's argument in one's own words. That is why deep argument analysis is also referred to as \emph{rational reconstruction} \citep[cf.][]{sep-carnap-suppD}. The reconstructed argument forms, together with details about its logical properties and about its relation to the source text, a \emph{comprehensive argumentative analysis} of a text. The latter can be seen as an interpretative hypothesis that is abductively inferred from a source text by means of an inference to the best explanation. Here is another example that illustrates how far a reconstruction may deviate from the original text that presents the argument \citep[adapted from][]{BrunBetzRaU2016}:

\noindent
\begin{tabular}{@{}c@{}c@{}c@{}}
\small{\textbf{source text}}&$\leadsto$&\small{\textbf{reconstructed argument}}\\
\begin{minipage}{.48\linewidth}
\fontsize{9}{10}\selectfont
So, the researcher's central dilemma exists in an especially acute form in psychology: either the animal is not like us, in which case there is no reason for performing the experiment; or else the animal is like us, in which case we ought not to perform on the animal an experiment that would be considered outrageous if performed on one of us. 
\end{minipage}&&
\begin{minipage}{.47\linewidth}
\fontsize{9}{10}\selectfont
(P1) If the animal is not like us, it is wrong to perform the experiment.

(P2) If the animal is like us, it is wrong to perform the experiment.

(C) \textsc{Thus} (with \emph{classical di\-lemma}): It is wrong to perform the experiment.
\end{minipage}
\end{tabular}\medskip

A compelling argumentative analysis yields (i) a rational argument that is (ii) closely related to the source text. Deep argument analysis is, accordingly, guided by a \textbf{dual goal} \citep[cf.][]{BrunBetzRaU2016}. An argument reconstruction should both be
\begin{itemize}
\setlength{\itemsep}{0mm}\setlength{\parskip}{0mm}
    \item[(i)] \textbf{systematically correct}, i.e., the reconstructed argument itself is, e.g., transparent, deductively valid, non-circular, or doesn't contain irrelevant premises; and  
    \item[(ii)] \textbf{exegetically adequate}, i.e., the reconstructed argument accounts for the original text, because, e.g., its premises merely reformulate parts of the text, or because its overall inferential structure can be traced within the source text.
\end{itemize}

The fact that there typically exists -- regarding a specific text -- a trade-off between these two goals is one major reason for the underdetermination of deep argument analysis and the plurality of legitimate reconstructions of a given text \citep[cf.][]{BrunBetzRaU2016}.

Against this background, we may finally define the problem of
\begin{description}
\item[Deep artificial argument analysis:] Describe, analyse and implement an effective computational system for deep argument analysis!
\end{description}






\subsection{Multi-angular Data}
\label{subsec:multi-angle}

\begin{figure}[tbp]
    \centering
    \input{figs/tikz_angles01}
    \vspace{-25pt}
    \caption{Relationships between dimensions of the multi-angular argumentative data.} 
    \label{fig:angles01}
\end{figure}
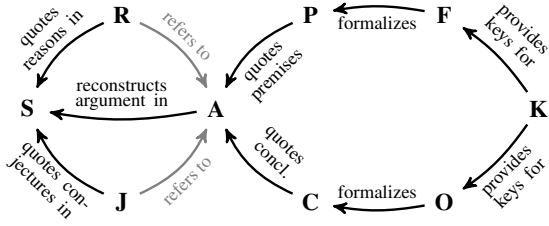

The DeepA2 framework is built upon a \emph{multi-angular} data structure \citep{TafjordClark2021GPQA} whose dimensions represent the essential components of a comprehensive argumentative analysis (see Section~\ref{subsec:problem}). Structured argumentative data is rendered as plain text \citep[cf.][]{Voigt2014}. The different data dimensions, which are related as shown in Figure~\ref{fig:angles01}, are (with an illustrating example):  

\begin{small}
\begin{description}
\setlength{\itemsep}{0mm}\setlength{\parskip}{0mm}
\item[argument source text (\small S)] \ \newline It is unethical to destroy human embryos. The basic argument supporting this claim just stresses that it is wrong to intentionally kill innocent human beings.
\item[verbatim reason statements in source text (\small R)]\ \newline it is wrong to intentionally kill innocent human beings (ref: (1))
\item[verbatim conjectures in the source text (\small J)]\ \newline It is unethical to destroy human embryos  (ref: (3))
\item[argument reconstruction (\small A)] {\ \newline (1) It is impermissible to kill innocent human beings.\newline
(2) The human embryo is an innocent human being.\newline
-- with hypothetical syllogism from (1) (2) --\newline 
(3) It is impermissible to kill the human embryo.}
\item[premises of the reconstructed argument (\small P)]\ \newline It is impermissible to kill innocent human beings $|$ The human embryo is an innocent human being
\item[final conclusion of reconstr.\ argument (\small C)]\ \newline It is impermissible to kill the human embryo
\item[formalizations of premises  (\small F)]\ \newline (x): F x $\rightarrow$ G x $|$ (x): H x $\rightarrow$ F x
\item[formalization of conclusion (\small O)]\ \newline (x): H x $\rightarrow$ G x
\item[keys for the formalizations' constants (\small K)]\ \newline F: innocent human being $|$ G: must not be killed $|$ H: human embryo
\end{description}
\end{small}

Each record in a DeepA2 dataset contains a source text plus a legitimate comprehensive argumentative analysis, which is, given underdetermination, not necessarily the only compelling reconstruction of the text; moreover, a dataset \emph{may} contain different records with one and the same source text analysed in several ways. So, for example, an alternative, equally legitimate argument reconstruction of the above source text (\textbf{\small S}) may read:

\begin{small}
\begin{description}
\setlength{\itemsep}{0mm}\setlength{\parskip}{0mm}
\item[argument reconstruction (\small A)] {\ \newline (1) If it is wrong to kill innocent human beings, then it is wrong to kill a human embryo.\newline
(2) It is wrong to kill innocent human beings.\newline
-- with modus ponens from (1) (2) --\newline 
(3) It is wrong to kill a human embryo.}
\end{description}
\end{small}

Beyond this structural and functional characterization, DeepA2 is agnostic about the nature and origin of the argumentative data. Synthetically generated, automatically retrieved, manually created datasets as well as translations of other databases are all compatible with the framework and can be used side by side.

\subsection{Generative Modes and Chains}
\label{subsec:generative_modes}

Given DeepA2's multi-dimensional data structure described in the previous section, a \textbf{generative mode} maps data from some input dimensions to a target dimension. For example, the mode \maa\ takes a source text (\textbf{\small S}) as input and outputs an argument reconstruction (\textbf{\small A}), the mode \mae\ reconstructs the argument (\textbf{\small A}) given the verbatim reasons (\textbf{\small R}) and conjectures (\textbf{\small J}). All in all, we define and investigate 21 different generative modes (see Appendix~\ref{app:training_setup}). Every mode represents a task on which a text-to-text model can be trained.   

By taking some mode's output as another mode's input, modes can be concatenated into \textbf{generative chains}. For example, the output of modes \mra\ and \mja\ (reasons and conjectures from source) can be fed into mode \mae\  to reconstruct an argument. Such generative chains allow us to emulate different strategies (heuristics) for analysing a given argumentative text (see Appendix~\ref{app:gen_chains} for technical details). 

Three generative chains which model distinct interpretative strategies, taking a source text (\textbf{\small S}) as sole input, are:

\begin{description}
\setlength{\itemsep}{0mm}\setlength{\parskip}{0mm}
\item[straight]\ \newline \maa\ \mra\ \mja \raggedright
\item[hermeneutic cycle]\ \newline \maa\ \mrc\ \mjc\ \mae  \raggedright
\item[logical streamlining]\ \newline \maa\ \mpa\ \mca\ \moa\ \mkb\ \mcb\ \maf\ \mrc\ \mjc \raggedright
\end{description}
While the chain \emph{straight}, where no output ever serves as input to another mode, represents a simple baseline, \emph{hermeneutic cycle} and \emph{logical streamlining} mimic prominent, equally-named methods in argument analysis \citep[cf.][]{BowllKemp2014,BrunBetzRaU2016}. One goes through a hermeneutic cycle, generally speaking, if one revisits a text in view of its previous interpretation, as, for example, in steps \mrc\ \mjc, where the source text (\textbf{\small S}) is re-interpreted (identifying reason statements and conjectures) given the previously reconstructed argument (\textbf{\small A}), so as to subsequently re-reconstruct the argument itself (step \mae). To logically streamline a reconstruction means to rephrase its conclusion or premises in order to make their logico-semantic structure more transparent. Such semantic clarification can be emulated by (i) formalizing a statement (e.g., \mca\ \moa\ \mkb) and (ii) using the keys (\textbf{\small K}) to retrieve the original statement from the generated logical formulas (such as in \mcb), from which the argument can be re-built (step \maf).

For evaluation, we append to each generative chain the following sub-chain that formalizes the reconstructed argument:   

\begin{description}
\item[formalization]\ \newline \mpa\ \mca\ \mfa\ \mob\ \mkc \raggedright\vspace{1mm}
\end{description}

A generative chain can be construed as hypergraph on the dimensions of DeepA2's multi-angular datasets, with each of its modes representing a directed hyper-edge. Summing up the number of input dimensions (except \textbf{\small S}) over all modes yields a simple graph centrality measure, which gauges a chain's sophistication. Thus, \emph{straight}, \emph{hermeneutic cycle} and \emph{logical streamlining} display a sophistication of 0, 4, and 11, respectively.


\subsection{Metrics}
\label{subsec:metrics}

As discussed in Section~\ref{subsec:problem}, an argument reconstruction should both be sound and make sense of the text to-be-interpreted. In line with the dual goal of argument analysis, we propose metrics both for the systematic correctness and for the exegetic adequacy of a given analysis. The following metrics measure the degree to which a given generated argument is \emph{systematically correct}:

\begin{description}
\setlength{\itemsep}{0mm}\setlength{\parskip}{0mm}
\item[\small SYS-PP] 1 if the argument is not a \emph{petitio principii} (i.e., if no premise is identical with its final conclusion), 0 otherwise;
\item[\small SYS-RP] 1 if the argument has no \emph{redundant premises} (i.e., if no premise occurs more than once), 0 otherwise;
\item[\small SYS-RC] 1 if the argument has no \emph{redundant conclusions} (i.e., if no conclusion -- intermediary or final -- occurs more than once), 0 otherwise; 
\item[\small SYS-US] 1 if all statements in the argument other than the final conclusion are explicitly \emph{used in an inference}, 0 otherwise; 
\item[\small SYS-SCH] ratio of sub-arguments which correctly instantiate the explicitly stated \emph{inference scheme} (e.g., hypothetical syllogism); 
\item[\small SYS-VAL] 1 if the argument is \emph{globally valid} (i.e., if the final conclusion deductively follows from the premises), 0 otherwise; 
\end{description}

All six systematic metrics can be computed automatically ({\small SYS-SCH} tries to parse the argument based on the inference schemes and templates used to construct the synthetic dataset in the first place; {\small SYS-VAL} passes the model-generated formalizations of premises and conclusion to a symbolic theorem prover \citep{de2008z3}; and the remaining metrics check for string identity). 

Whereas systematic metrics apply primarily to the generated argument (\textbf{\small A}), a reconstruction's interpretative adequacy will also depend on how reasons (\textbf{\small R}) and conjectures (\textbf{\small J}) coherently link the argument's components to the original text. As a first set of \emph{exegetic metrics}, we thus propose
\begin{description}
\setlength{\itemsep}{0mm}\setlength{\parskip}{0mm}
\item[\small EXE-MEQ] 1 if the reasons and conjectures are \emph{mutually exclusive verbatim quotes} from the source text, 0 otherwise;
\item[\small EXE-RSS] semantic similiarity \citep[BLEURT, see][]{sellam2020bleurt} of each reason statement and its counterpart premise in the reconstructed argument (if such exists, -1 otherwise);
\item[\small EXE-JSS] semantic similiarity (see {\small EXE-RSS}) of each conjecture statement and its counterpart in the reconstructed argument (if such exists, -1 otherwise).
\end{description}
Each source text presents (more or less faithfully) an underlying target argument, which in turn marks some of the text's statements as `target' reasons, others as `target' conjectures. The following two metrics assess the degree to which a comprehensive argumentative analysis correctly predicts (\textbf{\small R}, \textbf{\small J}) those target reasons and conjectures.
\begin{description}
\setlength{\itemsep}{0mm}\setlength{\parskip}{0mm}
\item[\small EXE-PPR] predictive performance (F1-score) for identifying (target) reason statements in the source text;
\item[\small EXE-PPJ] predictive performance (F1-score) for identifying (target) conjecture statements in the source text.
\end{description}
An argument's final conclusion may be implicit or explicit in a given text. The ability to fully exploit a text can be measured by verifying whether the reconstructed argument's final conclusion is implicit (= prediction) if and only if the target argument's one is. 
\begin{description}
\setlength{\itemsep}{0mm}\setlength{\parskip}{0mm}
\item[\small EXE-TE] text exploitation, as measured by ability (F1-score) to reconstruct arguments with explicit final conclusions (prediction) if and only if the target final conclusions are explicit.
\end{description}

\subsection{Models}
\label{subsec:models}

Any text-to-text language model is compatible with the proposed DeepA2 framework. We refer to models used within the framework as \textbf{ArgumentAnalyst}. In this study, we train and evaluate the transformer model T5 \citep{raffel2020exploring} with 770M parameters as implemented by \cite{wolf-etal-2020-transformers}.

\subsection{Limitations}

In the DeepA2 framework, arguments are reconstructed from relatively short and isolated texts, disregarding both the broader context of the argument and domain-specific background knowledge. This limits the framework, as presented here, in important ways: Implicit premises that are explicated in an argument reconstruction can neither be checked for plausibility nor for agreement with the author's broader convictions. In addition, the framework cannot assess an argument's dialectic function in a wider debate. It seems worthwhile to explore according extensions of the framework in future research.

\section{Datasets}
\label{sec:datasets}

For the experiments reported below, we synthetically create two artificial argument analysis corpora that comply with the DeepA2 framework (see also Appendix~\ref{app:aaac}): \textbf{\small AAAC01} and \textbf{\small AAAC02}. In addition, we translate the synthetic \emph{RuleTaker} \citep{Clark2020_TransSoftReas} and the manually compiled \emph{EntailmentBank} \citep{dalvi2021explaining} datasets into our framework. 

In argument analysis, one proceeds \emph{from} a source text \emph{to} its reconstruction. Creating the synthetic corpora, we reverse-engineer this process: 

\emph{Step 1.} We sample, first of all, a possibly complex argument (\textbf{\small A}) from a set of valid inference schemes. In doing so, we use a multi-step templating strategy \citep[inspired by][]{betz2020critical} to translate symbolic forms into natural language schemes (which were generated by local domain experts) and to substitute natural language terms for placeholders. Premises (\textbf{\small P}), conclusion (\textbf{\small C}) and their formalization (\textbf{\small F, O, K}) are side-products of such a construction of an argument.
    
\emph{Step 2.} Given the fully explicit argument (\textbf{\small A}), we compose a text (\textbf{\small S}) that presents the argument in a more or less transparent and faithful way. Such text creation involves: rendering the argument tree as a linear story, leaving out premises or conclusions (implicit premises and conclusions), inserting irrelevant material (distractors), using templates that obfuscate the logical form of a sentence, limiting the use of premise and conclusion indicators (such as ``therefore''), applying rule-based and automatic paraphrasing. In composing the argumentative text (\textbf{\small S}), we may record its reasons (\textbf{\small R}) and conjectures (\textbf{\small J}).


Given the synthetic and controlled nature of our dataset, which involved eliciting rule templates from a group of local domain experts,  all data is assumed to be correct by \emph{construction}. As an additional check of correctness on the logic of our examples, we ran a symbolic theorem prover \citep{de2008z3} over the argument formalizations to verify their validity. To ensure the fluency of the underlying language templates, all templates were hand verified by the authors.


Our two datasets {\small AAAC01} and {\small AAAC02} differ in the following ways:
\begin{enumerate}
\setlength{\itemsep}{0mm}\setlength{\parskip}{0mm}
    \item predicates and names are sampled from different, disjunct domains (texts are about, e.g., allergies and family relations versus, e.g., badminton and cooking) to test a model's robustness to lexical diversity \citep{RozenShwartzEtAl2019}; 
    \item similarly, {\small AAAC01} applies automatic paraphrasing \cite{Vamsi2021} to the final source text whereas {\small AAAC02} doesn't;
    \item {\small AAAC02} allows for imprecise renditions of logical formulas, while {\small AAAC01} sticks to plain formulations to test robustness to variations in description of rules. 
\end{enumerate}

Each dataset contains diverse texts and arguments. Broadly speaking, data records may differ in terms of properties of the argument (step 1 above) and properties of the argument's presentation (step 2). Along these two dimensions, we define five homogeneous subsets of the data:

\begin{description}
\setlength{\itemsep}{0mm}\setlength{\parskip}{0mm}
\item[simple inference:] arguments with a single inference step that neither involves negation nor compositional predicates;
\item[complex inference:] arguments with four inference steps that heavily rely on syntactically intricate schemes (e.g., transposition, or de Morgan);
\item[plain presentation:] all premises and conclusions are explicit in the source text which, in addition, contains no distractors;
\item[mutilated presentation:] at least two premises and one conclusion are implicit, while the text contains two distractors and explicitly states the final conclusion;
\item[C\&M:] the argument's inference is complex, plus the text contains at least two distractors.
\end{description}


The \emph{RuleTaker} and \emph{EntailmentBank} datasets contain multi-hop inference trees (\textbf{\small A}). To import these into the DeepA2 framework, we create source texts (\textbf{\small S}) for the given arguments by means of simple templates (such as ``\{\emph{theory}\} All this entails: \{\emph{hypothesis}\}'') and record reasons (\textbf{\small R}) and conjectures (\textbf{\small J}) on the fly. Unlike {\small AAAC} and \emph{EntailmentBank}, \emph{RuleTaker} \citep[as updated in][]{tafjord2020proofwriter} contains an equal share of arguments for which (i) the conclusion follows from the premises, (ii) the conclusion contradicts the premises, (iii) the conclusion is independent of the premises.

\section{Experiments and Results}
\label{sec:experiments}

\paragraph{As first and main experiment} we train our base model (see Section~\ref{subsec:models}) on the {\small AAAC01} corpus, and evaluate the resulting ArgumentAnalyst model out-of-domain on {\small AAAC02}. ArgumentAnalyst undergoes multi-task training on 21 generative modes, which are interpreted as sequence-to-sequence tasks (the training set-up is further described in Appendix~\ref{app:training_setup}).

The evaluation of ArgumentAnalyst on {\small AAAC02} proceeds in two steps: (1.) prediction: produces output in accordance with 16 different generative chains (Appendix~\ref{app:gen_chains}); (2.) metrics application: assesses the quality of the generated output by means of the systematic and exegetic metrics of the DeepA2 framework (see Section~\ref{subsec:metrics}).


\begin{table*}[htbp]
\begin{small}
\begin{tabularx}{\linewidth}{l *{12}{Y}}
\toprule
{} & \multicolumn{6}{c}{\emph{systematic metrics} (\textbf{\small SYS-*})} & \multicolumn{6}{c}{\emph{exegetic metrics} (\textbf{\small EXE-*})} \\ \cmidrule(r){2-7} \cmidrule(r){8-13}
chain&\textbf{\small PP}&\textbf{\small RP} &  \textbf{\small RC} &  \textbf{\small US} &  \textbf{\small SCH} &  \textbf{\small VAL} &  \textbf{\small MEQ} &  \textbf{\small RSS} &  \textbf{\small JSS} &  \textbf{\small PPR} &  \textbf{\small PPJ} &  \textbf{\small TE} \\
\midrule
straight &             .95 &                    .97 &                     .96 &                         .96 &                           .33 &            .73 &             .80 &                        -.08 &                       -.10 &         .93 &        .93 &               .63 \\
herm.\ cy. &             .95 &                    .98 &                     .95 &                         .93 &                           .31 &            .72 &             .82 &                         .16 &                        .12 &         .93 &        .92 &               .71 \\
logic.\ str. &             .95 &                    .97 &                     .96 &                         .95 &                           .32 &            .72 &             .82 &                         .11 &                       .00 &         .93 &        .92 &               .69 \\
pooling &             1.0 &                    1.0 &                     1.0 &                         1.0 &                           .73 &            1.0 &             1.0 &                         .26 &                        .29 &         .96 &        .96 &               .97 \\ 
\textit{oracle} &             \textit{1.0} &                    \textit{1.0} &                     \textit{1.0} &                         \textit{1.0} &                           \textit{1.0} &            \textit{1.0} &             \textit{1.0} &                         \textit{.30} &                        \textit{.37} &         \textit{1.0} &        \textit{1.0} &               \textit{1.0} \\
\bottomrule
\end{tabularx}
\end{small}
\caption{Performance of ArgumentAnalyst on the {\small AAAC02} data as measured by systematic and exegetic metrics. Rows display results for three illustrative generative chains (\emph{straight}, \emph{hermeneutic cycle}, \emph{logical streamlining}), for the item-wise best performing generative chain out of all 16 chains (\emph{pooling}), and for oracle performance (\emph{oracle}), which one obtains by applying the metrics to the target data itself.}
\label{table:main_results}
\end{table*}

Table~\ref{table:main_results} reports the ability of ArgumentAnalyst to generate systematically correct and exegetically adequate argument reconstructions. We obtain similar global results with the three chains \emph{straight}, \emph{hermeneutic cycle}, and \emph{logical streamlining}, whose generated reconstructions mainly differ in terms of internal coherence ({\small EXE-RSS}, {\small EXE-JSS}) and text exploitation ({\small EXE-TE}). However, the different generative chains complement each other, as shown by \emph{pooling}, which does not only outperform individual chains, but nearly attains oracle performance. 


\begin{table}[htbp]
\begin{small}
\begin{tabularx}{\linewidth}{l *{5}{Y}}
\toprule
{} & \multicolumn{2}{c}{\emph{ArgAn}\textsubscript{EB}}  & \multicolumn{2}{c}{\emph{ArgAn}\textsubscript{AAAC,EB}} & \emph{EntWr}\\ \cmidrule(l){2-3} \cmidrule(l){4-5}
steps &  straight &  herm.\ cycle &  straight &  herm.\ cycle & {}  \\
\midrule
1 & .863  & .866  & .816  & .871 & .951  \\
2 & .798  & .815  & .813  & .826 & .886  \\
3 & .812  & .815  & .826  & .806 & .858  \\
4 & .757  & .791  & .820  & .822 & .838  \\
$\geq$ 5 & .795  & .811  & .786  & .773 & .742  \\
any & .819  & .830  & .816  & .834 & .879  \\
\bottomrule\end{tabularx}
\caption{Predictive performance of ArgumentAnalyst ({\emph{ArgAn}\textsubscript{EB}}, \emph{ArgAn}\textsubscript{AAAC,EB}) and EntailmentWriter (\emph{EntWr}) for identifying reason statements in an input text (metric {\small SYS-PPR}) on the \emph{EntailmentBank task2} dataset.}
\label{table:ent_bank}
\end{small}
\end{table}

Moreover, ArgumentAnalyst produces much better reconstructions of simple inferences and plain presentations -- compared to complex inferences and mutilated presentations, i.e., difficult problems (cf.\ Table~\ref{table:main_subsets} in App.~\ref{app:add_results}). In addition, within one and the same subset, substantial differences show up between the three generative chains. Globally speaking, \emph{hermeneutic cycle} outperforms the other two chains for difficult problems.

\smallskip \noindent
\emph{Is {ArgumentAnalyst} capable of reliable self-evaluation?} We have \textbf{validated the logic metric} ({\small SYS-VAL}), which passes on a self-generated formalization of the reconstructed argument to a theorem prover, in three ways:  First of all, ArgumentAnalyst correctly recognizes \emph{target} arguments as valid (with accuracy 92.7\%), which has been verified by running the formalization subchain on target data. Secondly, virtually every generated argument with all-correct scheme instantiations (i.e., {\small SYS-SCH} $=1$) is also -- and correctly -- recognized as logically valid. Thirdly, a manual analysis (\textbf{human-in-the-loop}) of 100 generated arguments with incorrect scheme instantiation (i.e., {\small SYS-SCH} $<1$) reveals a high rate of false negatives: roughly one half of all inferences that are not automatically identified as an instantiation of the given scheme actually do correctly instantiate it. The accordingly \emph{adjusted} global ratio of correct scheme instantiations (Table~\ref{table:main_results}) equals roughly 0.65 (rather than 0.31--0.33), which is consistent with the ratio of logically valid arguments being 0.72--0.73.

\smallskip \noindent
\emph{Do reconstructed arguments exhibit basic semantic flaws?} Regarding the full dataset, ArgumentAnalyst produces nearly \textbf{flawless argument reconstructions}, committing basic errors (petitio, redundancy, unused statements) only very rarely (Table~\ref{table:main_results}). And even for very difficult problems, two thirds of all generated arguments display no basic flaw whatsoever (Table~\ref{table:main_subsets}, {\small SYS-PP \& SYS-RP \& SYS-RC \& SYS-US}).

\smallskip \noindent
\emph{Are reconstructed arguments logically valid?} Roughly 70\% of all arguments generated by one of the three chains are logically valid (Table~\ref{table:main_results}). More importantly, though, for virtually every source text in the dataset, there is at least one chain (out of 16) which reconstructs the text as a valid argument (\emph{pooling}). Given that logical validity can be automatically assessed, the \emph{pooled} system may thus \textbf{guarantee to yield a valid reconstruction}. Concerning different problem types (Table~\ref{table:main_subsets}), \emph{hermeneutic cycle} clearly outperforms the other chains as soon as the problem gets difficult. Additional analysis shows that ArgumentAnalyst can also \textbf{cope with underdetermination}, as 68\% of all generated arguments whose final conclusion differs ($\textrm{BLEU} \leq .8$) from the target argument's one -- i.e., arguments that are not reconstructed as expected given the target data -- are still logically valid.

\smallskip \noindent
\emph{Are the generated interpretations internally coherent?} The generative chain \emph{hermeneutic cycle} yields comprehensive argument reconstructions where premises (\textbf{\small P}) and conclusions (\textbf{\small C}) fit much better to detected reasons (\textbf{\small R}) and conjectures (\textbf{\small J}) than \emph{straight} or \emph{logical streamlining} ({\small EXE-RSS, EXE-JSS}). This holds globally (Table~\ref{table:main_results}), as well as for easy, and for difficult problems (Table~\ref{table:main_subsets}). Note that the \emph{oracle} baseline for metrics {\small EXE-RSS, EXE-JSS} is well below 1, which reflects the fact that source texts may present arguments in highly mutilated ways; it is nearly attained by \emph{pooling} the 16 different generative chains (Table~\ref{table:main_results}).    

\smallskip \noindent
\emph{Can ArgumentAnalyst detect reasons and conjectures, and fully exploit the text?}  The evaluation demonstrates that reason/conjecture detection on {\small AAAC02} is a relatively easy task ({\small EXE-PPR, EXE-PPJ}). In contrast, fully exploiting a text (i.e., generating an argument with implicit final conclusion if and only if the underlying target argument has an implicit final conclusion, {\small EXE-TE}) is seemingly more challenging (Table~\ref{table:main_results}). Again, \emph{hermeneutic cycle} achieves best text exploitation, performing, however, clearly below \emph{oracle} baseline -- which may simply reflect the degree of underdetermination in the {\small AAAC02} corpus.

\paragraph{In a second experiment} we train two models on the imported \emph{EntailmentBank} (\emph{task1} and \emph{task2}) dataset (see Section~\ref{sec:datasets}), namely: (1.) our base model (T5), which yields Argument\-Analyst\textsubscript{EB};
(2.) the ArgumentAnalyst model pretrained on {\small AAAC02} \citep[resulting in an intermediary pre-training set-up similar to][]{phang2018sentence,Geva2020InjectingNR}, which yields ArgumentAnalyst\textsubscript{AAAC,EB}.

Since the \emph{EntailmentBank} data doesn't contain formalizations, we can only train on 14 modes, which are interpreted as sequence-to-sequence tasks (see Appendix~\ref{app:training_setup}). We evaluate the models on \emph{task2} of \emph{EntailmentBank} only, which contains problems with a relatively large number of distractors, and proceed in two steps as before: prediction (with 11 different generative chains) and metrics application. \citet{dalvi2021explaining} report the ability of \emph{EntailmentWriter} (a fine-tuned T5-11b model) to correctly distinguish relevant premises of an argument from distractors in terms of a F1-score, which corresponds to our metric {\small EXE-PPR}. That's why the sole focus in this second experiment is on {\small EXE-PPR}. 

Table~\ref{table:ent_bank} describes the ability of ArgumentAnalyst models to correctly tell apart relevant premises from mere distractors in the \emph{EntailmentBank task2} dataset for two generative chains (\emph{straight}, which directly outputs reason statements, and \emph{hermeneutic cycle}, which tries to reconstruct the argument first and uses both source text and argument to identify reasons), and compares this with the performance of \emph{EntailmentWriter} \citep[scores from][]{dalvi2021explaining}. The results, shown separately for arguments with a specific number of inference steps, let us draw three conclusions:

    First, \emph{ArgumentAnalyst} outperforms \emph{EntailmentWriter} on difficult problems with more than 4 inference steps / sub-arguments.
    
    Second, using the sophisticated chain \emph{hermeneutic cycle} improves predictive performance compared to the simple \emph{straight} chain.

    Third, the chain \emph{hermeneutic cycle} (unlike \emph{straight}) generally benefits from intermediary pre-training on {\small AAAC} -- caveat: not so for arguments with more than 4 steps. This latter observation might be due to the fact that the {\small AAAC02} corpus, by construction, doesn't contain arguments with more than 4 steps, so that pre-training biases the model towards shorter arguments.  

\paragraph{In a third experiment} we explore the following hypothesis: 
\begin{description}
\item[Informative higher-order evidence.] The degree to which ArgumentAnalyst struggles in reconstructing a given argument (presented in the source text) as logically valid is a reliable indicator for whether the original argument is fallacious or not.
\end{description}
To test this hypothesis, we apply ArgumentAnalyst (trained on {\small AAAC02}, see above) to the \emph{RuleTaker} data as imported into the DeepA2 framework (see Section~\ref{sec:datasets}): ArgumentAnalyst produces -- by means of 13 generative chains -- comprehensive reconstructions, to which the systematic and exegetic metrics are applied. \emph{RuleTaker} contains an equal share of arguments whose conclusions follow from (label=valid), contradict (label=contradiction), or are independent of (label=neutral) the corresponding premises. Now, informative higher-order evidence would allow us to correctly predict these labels. And this is exactly what we observe: First, if reconstructions of one and the same source text which are independently generated with different chains agree (disagree), then the original argument tends to be valid (invalid). Second, by training simple classifiers on our argumentative metrics and further properties of the reconstructions, we robustly achieve a predictive accuracy 10\% above the random baseline. While this is far below the SOTA results of tailor-made RuleTaker \citep{Clark2020_TransSoftReas} and ProofWriter \citep{tafjord2020proofwriter} models on this data, our findings nonetheless confirm the above hypothesis.

\section{Conclusion}
\label{sec:conclusion}

In this paper, we have presented and implemented a multi-angular, modular framework for deep argument analysis (DeepA2). It allows for defining a large variety of generative modes by combining different dimensions of the data. These modes, in turn, can be concatenated into complex generative chains. ArgumentAnalyst -- a text-to-text model set up and trained within the DeepA2 framework -- yields plausible reconstructions of argumentative texts. Our empirical findings vindicate the overall framework and highlight the following \textbf{advantages of a multi-angular, modular design} in general:    
First of all, modular chains may emulate established, well-proven, typically piece-meal, scholarly techniques for text analysis (heuristics), which hence may provide \textbf{normative, methodological guidance} in setting up NLP systems. Secondly, by defining and implementing different modular chains, and investigating the plurality of generated solutions, one can systematically \textbf{explore the system's uncertainty as well as the tasks's underdetermination}. Thirdly, monitoring the system during modular computation yields diagnostically useful information (e.g., intermediary results) which not only describes the model's performance on the given problem, but which additionally allows us -- as \textbf{higher-order evidence} -- to characterize (e.g., classify) the original problem in the first place. Fourthly, breaking down a complex task into sub-tasks with intermediary results that can be further processed and re-combined helps to \textbf{overcome input size limitations} of neural language models. Fifthly, modular generation with meaningful modes allows users to follow the system, comprehend generated solutions, verify sub-steps and detect errors -- the NLP system becomes a \textbf{transparent, explainable AI} \citep{Miller2019ExplanationIA}. Finally, modular NLP systems as described by DeepA2 may be connected to a user-interface which promises \textbf{fine-grained interactive control} of modular generations and seamless cognitive cooperation of AI and human experts in analysing texts.



%% file: figs/basic_design_tacl.tex
\setstretch{0.7}
\begin{tikzpicture}[>=stealth',line width=1pt, node distance=45pt]
\node[roundnode,fill=colbrew3] (input1) {\fontsize{8}{9}\selectfont\texttt{conjectures: source: Socrates is mortal because every human is.}};
\node[roundnode,fill=colbrew1,below of=input1] (input2) {\fontsize{8}{9}\selectfont \texttt{argdown: source: Socrates is mortal because every human is. conjectures: Socrates is mortal}};
\node[roundnode,fill=colbrew6,below of=input2, node distance=45pt] (input3) {\fontsize{8}{9}\selectfont \texttt{formalize: premises: Socrates is human | If someone is human, then they are mortal}};
\node[roundnode2,right of=input2,node distance=150pt, align=center] (processor) { \textbf{Argument-}\\\textbf{Analyst}}
   edge [->,pre,in=0,out=170] node {} (input1)
   edge [->,pre,in=0,out=180] node {} (input2)
   edge [->,pre,in=0,out=190] node {} (input3);
\node[roundnode,fill=colbrew1,right of=processor,node distance=150pt] (out2) {\fontsize{8}{9}\selectfont \texttt{(1) Socrates is human.\newline
(2) If someone is human, then they are mortal.\newline
----\newline
(3) Socrates is mortal.}}
   edge [->,pre,in=0,out=180] node {} (processor);
\node[roundnode,fill=colbrew3,above of=out2] (out1) {\fontsize{8}{9}\selectfont \texttt{Socrates is mortal}}
   edge [->,pre,in=10,out=180] node {} (processor);
\node[roundnode,fill=colbrew6,below of=out2] (out3) {\fontsize{8}{9}\selectfont \texttt{F a | (x): F x -> G x}}
   edge [->,pre,in=350,out=180] node {} (processor);

\end{tikzpicture}

%% file: figs/tikz_angles01.tex
\begin{small}\centering
\begin{tikzpicture}[>=stealth',line width=.9pt,auto]
\node[vertex2,label=center:\(\textbf{S}\)] (S) {};
\node[vertex2,above right of=S,label=center:\(\textbf{R}\)] (R) {}
   edge [post,bend right=20] node[align=center,sloped,text width=40pt,execute at begin node=\setlength{\baselineskip}{1ex}] {\scriptsize quotes reasons in} (S);
\node[vertex2,below right of=S,label=center:\(\textbf{J}\)] (J) {}
   edge [post,bend left=20] node[sloped,align=center,yshift=-18pt,text width=40pt,execute at begin node=\setlength{\baselineskip}{1ex}] {\scriptsize quotes con- jectures in} (S);
\node[vertex2,below right of=R,label=center:\(\textbf{A}\)] (A) {}
   edge [post,bend left=10] node[align=center,sloped,text width=40pt,execute at begin node=\setlength{\baselineskip}{1ex}] {\scriptsize reconstructs argument in} (S)
   edge [pre,bend right=20,color=gray] node[align=center,sloped,xshift=-2pt] {\scriptsize refers to} (R)
   edge [pre,bend left=20,color=gray] node[align=center,sloped,yshift=-11pt,xshift=-2pt] {\scriptsize refers to} (J);
\node[vertex2,above right of=A,label=center:\(\textbf{P}\)] (P) {}
   edge [post,bend right=20] node[align=center,sloped,xshift=0pt,yshift=-20pt,text width=40pt,execute at begin node=\setlength{\baselineskip}{1ex}] {\scriptsize quotes premises} (A);
\node[vertex2,below right of=A,label=center:\(\textbf{C}\)] (C) {}
   edge [post,bend left=20] node[align=center,sloped,xshift=3pt,yshift=2pt,text width=30pt,execute at begin node=\setlength{\baselineskip}{1ex}] {\scriptsize quotes concl.} (A);
\node[vertex2,right of=P,label=center:\(\textbf{F}\)] (F) {}
   edge [post,bend right=10] node[align=center,sloped,yshift=-12pt] {\scriptsize formalizes} (P);
\node[vertex2,right of=C,label=center:\(\textbf{O}\)] (O) {}
   edge [post,bend left=10] node[align=center,sloped,yshift=1pt] {\scriptsize formalizes} (C);
\node[vertex2,below right of=F,label=center:\(\textbf{K}\)] (K) {}
   edge [post,bend right=10] node[align=center,sloped,text width=40pt,execute at begin node=\setlength{\baselineskip}{1ex}] {\scriptsize provides keys for} (F)
   edge [post,bend left=10] node[align=center,sloped,yshift=-19pt,text width=40pt,execute at begin node=\setlength{\baselineskip}{1ex}] {\scriptsize provides keys for} (O);

\end{tikzpicture}
\end{small}

%% file: appendix.tex
\section{Synthetic Argument Data}
\label{app:aaac}

The {\small AAAC} datasets used in this study are publicly available via Huggingface's Hub -- {\small \url{https://huggingface.co/datasets/debatelab/aaac}} -- where the construction of the datasets is documented meticulously.

A synthetically generated {\small AAAC} record, which nicely illustrates the underdetermination of argument reconstruction, with two implicit premises, one distracting statement and a simple (one-step) argument (formatted as presented to the model): 

\begin{footnotesize}\ttfamily
\noindent\textit{source:} It is not the case that Tracy is not an admirer of Fullerton and Tracy has seen La Habra. Plus, if someone loves Chico, then they haven't visited Monterey, owing to the fact that loving Laguna Beach is sufficient for not having visited Monterey.

\noindent\textit{reasons:} loving Laguna Beach is sufficient for not having visited Monterey (ref: (2))

\noindent\textit{conjectures:} if someone loves Chico, then they haven't visited Monterey (ref: (4))


\noindent\textit{argdown:}\newline
(1) If someone is an admirer of Chico, then they are an admirer of Laguna Beach or a visitor of Stockton.\newline
(2) If someone admires Laguna Beach, then they haven't visited Monterey.\newline
(3) If someone has visited Stockton, then they haven't visited Monterey.\newline
--\newline
with generalized dilemma (neg variant) from (1) (2) (3)\newline
--\newline
(4) If someone admires Chico, then they haven't visited Monterey.

\noindent\textit{premises:} 
If someone is an admirer of Chico, then they are an admirer of Laguna Beach or a visitor of Stockton. (ref: (1)) |
If someone admires Laguna Beach, then they haven't visited Monterey. (ref: (2)) |
If someone has visited Stockton, then they haven't visited Monterey. (ref: (3))

\noindent\textit{conclusion:} 
If someone admires Chico, then they haven't visited Monterey. (ref: (4)) 

\noindent\textit{premises\_form:} 
(x): Fx -> (G x v H x) (ref: (1)) |
(x): G x -> not I x (ref: (2)) |
(x): H x -> not I x (ref: (3))

\noindent\textit{conclusion\_form:} 
(x): F x -> not I x (ref: (4)) 

\noindent\textit{keys:} 
F: admirer of Chico |
G: admirer of Laguna Beach |
H: visitor of Stockton |
I: visitor of Monterey

\end{footnotesize}

\section{Training Set-up}
\label{app:training_setup}

By interpreting a generative mode as a sequence-to-sequence task, we may translate a multi-angular DeepA2 dataset (e.g., {\small AAAC01}) into a multi-task sequence-to-sequence format, on which a sequence-to-sequence model can be trained. For each record in the multi-angular DeepA2 dataset, we randomly sample 14 modes in accordance with the weights provided in Table~\ref{table:all_generative_modes} and add, for each mode, a corresponding sequence-to-sequence record to the training data. This results, for {\small AAAC01}, in a sequence-to-sequence training dataset with $14\times 16.000$ records.

\begin{table}[tb]
    \centering
    \begin{small}
    \begin{tabularx}{\linewidth}{@{}p{0.20\linewidth}@{}Y@{}Y|p{0.17\linewidth}@{}Y@{}Y|p{0.23\linewidth}@{}Y@{}Y@{}}
    \toprule
    mode & w\textsubscript 1 & w\textsubscript{2} & mode & w\textsubscript 1 & w\textsubscript{2} & mode & w\textsubscript 1 & w\textsubscript{2} \\ 
    \midrule
    \maa  & 1. & 1. & \mra  & 1. & 1. & \mfa  & .7 & -- \\
    \mab  & 1. & 1. & \mrb  & 1. & 1. & \mfb  & .7 & -- \\
    \mac  & 1. & 1. & \mrc  & 1. & 1. & \moa  & .7 & -- \\
    \mad  & 1. & 1. & \mja  & 1. & 1. & \mob  & .7 & -- \\
    \mae  & 1. & 1. & \mjb  & 1. & 1. & \mka  & .7 & -- \\
    \maf  & 1. & 1. & \mjc  & 1. & 1. & \mkb  & .7 & -- \\
    \mpa  & .2 & .2 & \mca  & .2 & .2 & \mkc  & .7 & -- \\
    \mpb  & .7 & -- & \mcb  & .7 & -- &       &    &    \\
    \bottomrule
    \end{tabularx}
    \end{small}
    \caption{21 generative modes with corresponding weights in {\small AAAC} (w\textsubscript 1) and \emph{EntailmentBank} (w\textsubscript 2) training data.}
    \label{table:all_generative_modes}
\end{table}

Our models (base model T5-large with 770M parameters, and pretrained ArgumentAnalyst) are trained with batch-size 2 and learning rate 0.00001. For {\small AAAC01}, eval loss starts to increase at epoch 8; with \emph{EntailmentBank} data, eval loss increases from epoch 2 onwards.

\section{Iterative Prediction with Generative Chains}
\label{app:gen_chains}

Generative chains are implemented with a dynamic dictionary (9 keys, corresp.\ to the dimensions of DeepA2 data), which is initialized with the source text, provides input for the generative modes, and is updated after each generative step with the mode's generated output. Output is generated with beam search decoding and beam width 2.

\begin{table}[tb]
    \centering
    \begin{small}
    \begin{tabularx}{\linewidth}{@{}lp{0.68\linewidth}c@{\hspace{3pt}}c@{}}
    \toprule
    \# & {mode sequence}  & len. & soph. \\ 
    \midrule
    \textbf{1} &  \maa\ \mra\ \mja  & 3 & 0 \smallskip\\
    2 &    \mja\ \mra\ \mac  & 3 & 1 \smallskip\\
    3 &    \mja\ \mra\ \mab  & 3 & 1 \smallskip\\
    4 &    \mja\ \mra\ \mae  & 3 & 2 \smallskip\\
    5 &    \mja\ \mrb\ \mae  & 3 & 3 \smallskip\\
    6 &    \mja\ \mrb\ \mad  & 3 & 3 \smallskip\\
    7 &    \mra\ \mjb\ \mae  & 3 & 3 \smallskip\\
    8 &    \mra\ \mjb\ \mad  & 3 & 3 \smallskip\\
    \textbf{9} & \maa\ \mrc\ \mjc\ \mae  & 4 & 4 \smallskip\\
    10 &   \maa\ \mrc\ \mjc\ \mad  & 4 & 4 \smallskip\\
    11 & \parbox[t]{\linewidth}{ \raggedright   \maa\ \mrc\ \mjc\ \mad\ \mrc\ \mjc\ \mad } & 7 & 8 \smallskip\\
    12 &\parbox[t]{\linewidth}{ \raggedright   \maa\ \mpa\ \mca\ \mfa\ \mka\ \mpb\ \maf\ \mrc\ \mjc}  & 9 & 11 \smallskip\\
    \textbf{13} &\parbox[t]{\linewidth}{ \raggedright   \maa\ \mpa\ \mca\ \moa\ \mkb\ \mcb\ \maf\ \mrc\ \mjc}  & 9 & 11 \smallskip\\
    14 &\parbox[t]{\linewidth}{  \raggedright \maa\ \mpa\ \mca\ \moa\ \mkb\ \mcb\ \maf\ \mpa\ \mca\ \mfa\ \mka\ \mpb\ \maf\ \mrc\ \mjc }\vspace{1pt} & 15 & 20 \smallskip\\
    15 &\parbox[t]{\linewidth}{ \raggedright \maa\ \mpa\ \mca\ \mfa\ \mob\ \mkc\ \mpb\ \mcb\ \maf\ \mrc\ \mjc } & 11 & 18 \smallskip\\
    16 & \parbox[t]{\linewidth}{  \raggedright \maa\ \mpa\ \mca\ \mfa\ \mob\ \mfb\ \mkc\ \mpb\ \mcb\ \maf\ \mrc\ \mjc } & 12 & 21 \\
    \bottomrule
    \end{tabularx}
    \end{small}
    \caption{16 generative chains (without final formalization sub-sequences) evaluated in this study. The illustrative chains highlighted in the main paper are \#1 (straight), \#9 (hermeneutic cycle), and \#13 (logical streamlining).}
    \label{table:all_generative_chains_app}
\end{table}

Table~\ref{table:all_generative_chains_app} displays all generative chains we resort to in this study, all of which are used in the \textit{first experiment}. The \textit{second experiment} makes use of chains 1--11. The \textit{third experiment} deploys chains 1--13.   

\section{Additional Results}
\label{app:add_results}

Table~\ref{table:main_subsets} assesses ArgumentAnalyst's reconstructions on specific subsets of the {\small AAAC02} dataset (defined in Section~\ref{sec:datasets}) for three representative generative chains. 

\begin{table}[tb]
\begin{small}
\begin{tabularx}{\linewidth}{l *{5}{Y}}
\toprule
{} & \multicolumn{2}{c}{\emph{inference}}  & \multicolumn{2}{c}{\emph{presentation}} \\ \cmidrule(l){2-3} \cmidrule(l){4-5}
{} &  {\small simple} &  {\small compl.} &  {\small plain} &  {\small mutil.} &  {\small C\&M}  \\
chain &  {\scriptsize N=1274} &   {\scriptsize N=180} &   {\scriptsize N=330} &   {\scriptsize N=114} &  {\scriptsize N=70}  \\
\midrule
\multicolumn{6}{c}{\textbf{\scriptsize SYS-PP \& SYS-RP \& SYS-RC \& SYS-US}} \\
straight &              .95 &               .72 &                .98 &                    .61 &               .69 \\
herm.\ c. &              .94 &               .68 &                .96 &                    .67 &               .61 \\
log.\ str. &              .95 &               .68 &                .98 &                    .64 &               .61 \\
\midrule
\multicolumn{6}{c}{\textbf{\scriptsize SYS-VAL}} \\
straight &              .84 &               .48 &                .88 &                    .40 &               .34 \\
herm.\ c. &              .83 &               .56 &                .84 &                    .49 &               .50 \\
log.\ str. &              .82 &               .47 &                .86 &                    .46 &               .37 \\\midrule
\multicolumn{6}{c}{\textbf{\scriptsize EXE-RSS}} \\
straight &              .03 &              -.25 &                .05 &                   -.31 &              -.30 \\
herm.\ c. &              .20 &               .08 &                .15 &                    .08 &               .11 \\
log.\ str. &              .17 &              -.01 &                .13 &                    .01 &              -.06 \\
\midrule
\multicolumn{6}{c}{\textbf{\scriptsize EXE-JSS}} \\
straight &              .06 &              -.32 &                .10 &                   -.37 &              -.37 \\
herm.\ c. &              .23 &              -.06 &                .21 &                   -.03 &              -.21 \\
log.\ str. &              .13 &              -.26 &                .07 &                   -.26 &              -.40 \\
\bottomrule\end{tabularx}
\caption{Performance of ArgumentAnalyst on specific subsets (columns) of the {\small AAAC02} data as measured by selected systematic and exegetic metrics (sub-tables). Rows display results for three illustrative generative chains (\emph{straight}, \emph{hermeneutic cycle}, \emph{logical streamlining}).}
\label{table:main_subsets}
\end{small}
\end{table}

Table~\ref{table:main_results_app} details the performance of ArgumentAnalyst on the entire {\small AAAC02} dataset as measured by tailor-made argumentative metrics. Table~\ref{table:main_results_app_oos} shows the corresponding performance on out-of -sample eval data {\small AAAC01}.   

\begin{table*}
\begin{small}
\begin{tabularx}{\linewidth}{l *{12}{Y}}
\toprule
{} & \multicolumn{6}{c}{\emph{systematic metrics} (\textbf{\scriptsize SYS-*})} & \multicolumn{6}{c}{\emph{exegetic metrics} (\textbf{\scriptsize EXE-*})} \\ \cmidrule(r){2-7} \cmidrule(r){8-13}
chain &  \textbf{\scriptsize PP} &  \textbf{\scriptsize RP} &  \textbf{\scriptsize RC} &  \textbf{\scriptsize US} &  \textbf{\scriptsize SCH} &  \textbf{\scriptsize VAL} &  \textbf{\scriptsize MEQ} &  \textbf{\scriptsize RSS} &  \textbf{\scriptsize JSS} &  \textbf{\scriptsize PPR} &  \textbf{\scriptsize PPJ} &  \textbf{\scriptsize TE} \\
\midrule
\#1  &             0.95 &                    0.97 &                     0.96 &                         0.96 &                           0.33 &            0.73 &             0.80 &                        -0.08 &                       -0.10 &         0.93 &        0.93 &               0.63 \\
\#2  &             0.95 &                    0.97 &                     0.94 &                         0.94 &                           0.33 &            0.71 &             0.80 &                        -0.09 &                        0.04 &         0.93 &        0.93 &               0.67 \\
\#3  &             0.95 &                    0.98 &                     0.95 &                         0.93 &                           0.31 &            0.70 &             0.80 &                         0.10 &                       -0.11 &         0.93 &        0.93 &               0.62 \\
\#4  &             0.94 &                    0.97 &                     0.94 &                         0.92 &                           0.30 &            0.70 &             0.80 &                         0.12 &                       -0.00 &         0.93 &        0.93 &               0.66 \\
\#5  &             0.94 &                    0.97 &                     0.95 &                         0.91 &                           0.30 &            0.70 &             0.83 &                         0.13 &                        0.05 &         0.94 &        0.93 &               0.69 \\
\#6  &             0.94 &                    0.97 &                     0.95 &                         0.93 &                           0.31 &            0.70 &             0.83 &                         0.10 &                        0.03 &         0.94 &        0.93 &               0.67 \\
\#7  &             0.93 &                    0.97 &                     0.95 &                         0.92 &                           0.29 &            0.70 &             0.83 &                         0.13 &                        0.05 &         0.93 &        0.92 &               0.68 \\
\#8  &             0.94 &                    0.97 &                     0.95 &                         0.93 &                           0.30 &            0.69 &             0.83 &                         0.10 &                        0.02 &         0.93 &        0.92 &               0.67 \\
\#9  &             0.95 &                    0.98 &                     0.95 &                         0.93 &                           0.31 &            0.72 &             0.82 &                         0.16 &                        0.12 &         0.93 &        0.92 &               0.71 \\
\#10  &             0.96 &                    0.98 &                     0.96 &                         0.94 &                           0.32 &            0.71 &             0.82 &                         0.14 &                        0.09 &         0.93 &        0.92 &               0.69 \\
\#11 &             0.96 &                    0.98 &                     0.96 &                         0.93 &                           0.32 &            0.71 &             0.82 &                         0.15 &                        0.11 &         0.93 &        0.92 &               0.71 \\
\#12 &             0.93 &                    0.95 &                     0.94 &                         0.94 &                           0.32 &            0.71 &             0.81 &                        -0.17 &                       -0.08 &         0.93 &        0.92 &               0.68 \\
\#13 &             0.95 &                    0.97 &                     0.96 &                         0.95 &                           0.32 &            0.72 &             0.82 &                         0.11 &                       -0.00 &         0.93 &        0.92 &               0.69 \\
\#14 &             0.93 &                    0.95 &                     0.94 &                         0.94 &                           0.32 &            0.70 &             0.81 &                        -0.18 &                       -0.14 &         0.93 &        0.92 &               0.66 \\
\#15 &             0.92 &                    0.96 &                     0.94 &                         0.95 &                           0.33 &            0.71 &             0.81 &                        -0.20 &                       -0.19 &         0.93 &        0.92 &               0.65 \\
\#16 &             0.92 &                    0.96 &                     0.94 &                         0.94 &                           0.33 &            0.72 &             0.81 &                        -0.20 &                       -0.19 &         0.93 &        0.92 &               0.65 \\
\bottomrule
\end{tabularx}
\end{small}
\caption{Performance of ArgumentAnalyst for systematic and exegetic metrics on the entire OOD eval data ({\small AAAC02}). Rows display mean results for each of the 16 generative chains.}
\label{table:main_results_app}
\end{table*}

\begin{table*}
\begin{small}
\begin{tabularx}{\linewidth}{l *{12}{Y}}
\toprule
{} & \multicolumn{6}{c}{\emph{systematic metrics} (\textbf{\scriptsize SYS-*})} & \multicolumn{6}{c}{\emph{exegetic metrics} (\textbf{\scriptsize EXE-*})} \\ \cmidrule(r){2-7} \cmidrule(r){8-13}
chain &  \textbf{\scriptsize PP} &  \textbf{\scriptsize RP} &  \textbf{\scriptsize RC} &  \textbf{\scriptsize US} &  \textbf{\scriptsize SCH} &  \textbf{\scriptsize VAL} &  \textbf{\scriptsize MEQ} &  \textbf{\scriptsize RSS} &  \textbf{\scriptsize JSS} &  \textbf{\scriptsize PPR} &  \textbf{\scriptsize PPJ} &  \textbf{\scriptsize TE} \\
\midrule
\#1  &             0.97 &                    0.98 &                     0.97 &                         0.98 &                           0.61 &            0.87 &             0.78 &                         0.08 &                        0.13 &         0.95 &        0.95 &               0.64 \\
\#2  &             0.97 &                    0.98 &                     0.96 &                         0.97 &                           0.60 &            0.87 &             0.78 &                         0.09 &                        0.24 &         0.95 &        0.95 &               0.68 \\
\#3  &             0.96 &                    0.98 &                     0.96 &                         0.97 &                           0.58 &            0.86 &             0.78 &                         0.26 &                        0.12 &         0.95 &        0.95 &               0.64 \\
\#4  &             0.95 &                    0.98 &                     0.95 &                         0.96 &                           0.57 &            0.85 &             0.78 &                         0.26 &                        0.20 &         0.95 &        0.95 &               0.67 \\
\#5  &             0.96 &                    0.98 &                     0.95 &                         0.96 &                           0.57 &            0.84 &             0.80 &                         0.27 &                        0.27 &         0.96 &        0.95 &               0.70 \\
\#6  &             0.97 &                    0.98 &                     0.96 &                         0.96 &                           0.58 &            0.84 &             0.80 &                         0.26 &                        0.24 &         0.96 &        0.95 &               0.69 \\
\#7  &             0.95 &                    0.98 &                     0.96 &                         0.96 &                           0.57 &            0.86 &             0.79 &                         0.27 &                        0.26 &         0.95 &        0.94 &               0.71 \\
\#8  &             0.96 &                    0.98 &                     0.96 &                         0.96 &                           0.57 &            0.85 &             0.79 &                         0.26 &                        0.25 &         0.95 &        0.94 &               0.70 \\
\#9  &             0.97 &                    0.99 &                     0.97 &                         0.97 &                           0.59 &            0.88 &             0.79 &                         0.31 &                        0.36 &         0.96 &        0.95 &               0.78 \\
\#10  &             0.97 &                    0.99 &                     0.97 &                         0.97 &                           0.60 &            0.87 &             0.79 &                         0.30 &                        0.34 &         0.96 &        0.95 &               0.77 \\
\#11 &             0.97 &                    0.99 &                     0.97 &                         0.97 &                           0.60 &            0.87 &             0.79 &                         0.31 &                        0.35 &         0.96 &        0.95 &               0.77 \\
\#12 &             0.95 &                    0.97 &                     0.95 &                         0.96 &                           0.54 &            0.84 &             0.79 &                         0.17 &                        0.25 &         0.96 &        0.94 &               0.75 \\
\#13 &             0.97 &                    0.99 &                     0.97 &                         0.97 &                           0.61 &            0.87 &             0.79 &                         0.29 &                        0.32 &         0.96 &        0.95 &               0.76 \\
\#14 &             0.95 &                    0.97 &                     0.95 &                         0.96 &                           0.54 &            0.84 &             0.79 &                         0.16 &                        0.24 &         0.96 &        0.94 &               0.74 \\
\#15 &             0.94 &                    0.97 &                     0.95 &                         0.96 &                           0.54 &            0.85 &             0.79 &                         0.15 &                        0.18 &         0.96 &        0.95 &               0.73 \\
\#16 &             0.94 &                    0.97 &                     0.95 &                         0.95 &                           0.54 &            0.85 &             0.79 &                         0.15 &                        0.19 &         0.96 &        0.95 &               0.73 \\
\bottomrule
\end{tabularx}
\end{small}
\caption{Performance of ArgumentAnalyst for systematic and exegetic metrics on the entire OOS eval data ({\small AAAC01}). Rows display mean results for each of the 16 generative chains.}
\label{table:main_results_app_oos}
\end{table*}

Distinguishing four mutually exclusive subsets of {\small AAAC02}, Tables~\ref{table_main_subsets1}--\ref{table_main_subsets4} detail the the quality of ArgumentAnalyst's reconstruction for easy and difficult problems. Tables~\ref{table_main_subsets_oos1}--\ref{table_main_subsets_oos4} present the corresponding out-of-sample performance on the equally partitioned {\small AAAC01} dataset (eval split).

\begin{table}
\begin{small}
\begin{tabularx}{\linewidth}{l *{5}{Y}}
\toprule
{} & \multicolumn{2}{c}{\emph{inference}}  & \multicolumn{2}{c}{\emph{presentation}} \\ \cmidrule(l){2-3} \cmidrule(l){4-5}
chain &  simple &  complex &  plain &  mutilat. & C\&M  \\
\midrule
\multicolumn{6}{c}{\textbf{\scriptsize SYS-PP $\&$ SYS-RP $\&$ SYS-RC $\&$ SYS-US}} \\
\#1  &              0.95 &               0.72 &                0.98 &                    0.61 &               0.69 \\
\#2  &              0.93 &               0.66 &                0.96 &                    0.59 &               0.60 \\
\#3  &              0.92 &               0.69 &                0.96 &                    0.68 &               0.73 \\
\#4  &              0.92 &               0.66 &                0.95 &                    0.69 &               0.60 \\
\#5  &              0.92 &               0.68 &                0.95 &                    0.59 &               0.61 \\
\#6  &              0.93 &               0.66 &                0.97 &                    0.68 &               0.59 \\
\#7  &              0.92 &               0.67 &                0.96 &                    0.62 &               0.64 \\
\#8  &              0.92 &               0.66 &                0.95 &                    0.64 &               0.66 \\
\#9  &              0.94 &               0.68 &                0.96 &                    0.67 &               0.61 \\
\#10  &              0.94 &               0.73 &                0.98 &                    0.68 &               0.77 \\
\#11 &              0.94 &               0.69 &                0.98 &                    0.66 &               0.73 \\
\#12 &              0.93 &               0.60 &                0.95 &                    0.57 &               0.50 \\
\#13 &              0.95 &               0.68 &                0.98 &                    0.64 &               0.61 \\
\#14 &              0.92 &               0.57 &                0.93 &                    0.58 &               0.49 \\
\#15 &              0.92 &               0.66 &                0.95 &                    0.59 &               0.56 \\
\#16 &              0.92 &               0.64 &                0.95 &                    0.56 &               0.60 \\
\bottomrule\end{tabularx}
\caption{Performance of ArgumentAnalyst for selected systematic metric (\textbf{\scriptsize SYS-PP $\&$ SYS-RP $\&$ SYS-RC $\&$ SYS-US}) on specific subsets (columns) of the OOD eval data.}
\label{table_main_subsets1}
\end{small}
\end{table}

\begin{table}
\begin{small}
\begin{tabularx}{\linewidth}{l *{5}{Y}}
\toprule
{} & \multicolumn{2}{c}{\emph{inference}}  & \multicolumn{2}{c}{\emph{presentation}} \\ \cmidrule(l){2-3} \cmidrule(l){4-5}
chain &  simple &  complex &  plain &  mutilat. & C\&M  \\
\midrule
\multicolumn{6}{c}{\textbf{\scriptsize SYS-VAL}} \\
\#1  &              0.84 &               0.48 &                0.88 &                    0.40 &               0.34 \\
\#2  &              0.82 &               0.54 &                0.84 &                    0.47 &               0.46 \\
\#3  &              0.82 &               0.44 &                0.87 &                    0.39 &               0.36 \\
\#4  &              0.81 &               0.48 &                0.83 &                    0.44 &               0.43 \\
\#5  &              0.82 &               0.44 &                0.85 &                    0.45 &               0.37 \\
\#6  &              0.81 &               0.46 &                0.85 &                    0.42 &               0.41 \\
\#7  &              0.83 &               0.44 &                0.82 &                    0.46 &               0.49 \\
\#8  &              0.80 &               0.44 &                0.83 &                    0.40 &               0.40 \\
\#9  &              0.83 &               0.56 &                0.84 &                    0.49 &               0.50 \\
\#10  &              0.82 &               0.50 &                0.85 &                    0.46 &               0.43 \\
\#11 &              0.82 &               0.48 &                0.84 &                    0.46 &               0.41 \\
\#12 &              0.81 &               0.47 &                0.84 &                    0.42 &               0.37 \\
\#13 &              0.82 &               0.47 &                0.86 &                    0.46 &               0.37 \\
\#14 &              0.80 &               0.48 &                0.82 &                    0.41 &               0.40 \\
\#15 &              0.82 &               0.45 &                0.84 &                    0.50 &               0.33 \\
\#16 &              0.83 &               0.52 &                0.85 &                    0.46 &               0.43 \\
\bottomrule\end{tabularx}
\caption{Performance of ArgumentAnalyst for selected systematic metric (\textbf{\scriptsize SYS-VAL}) on specific subsets (columns) of the OOD eval data.}
\label{table_main_subsets2}
\end{small}
\end{table}

\begin{table}
\begin{small}
\begin{tabularx}{\linewidth}{l *{5}{Y}}
\toprule
{} & \multicolumn{2}{c}{\emph{inference}}  & \multicolumn{2}{c}{\emph{presentation}} \\ \cmidrule(l){2-3} \cmidrule(l){4-5}
chain &  simple &  complex &  plain &  mutilat. & C\&M  \\
\midrule
\multicolumn{6}{c}{\textbf{\scriptsize EXE-RSS}} \\
\#1  &              0.03 &              -0.25 &                0.05 &                   -0.31 &              -0.30 \\
\#2  &              0.02 &              -0.27 &                0.07 &                   -0.33 &              -0.31 \\
\#3  &              0.15 &              -0.03 &                0.12 &                   -0.01 &              -0.06 \\
\#4  &              0.16 &               0.01 &                0.12 &                   -0.01 &               0.04 \\
\#5  &              0.18 &               0.04 &                0.13 &                    0.04 &               0.06 \\
\#6  &              0.17 &              -0.04 &                0.12 &                   -0.02 &              -0.09 \\
\#7  &              0.18 &               0.05 &                0.14 &                    0.03 &               0.08 \\
\#8  &              0.16 &              -0.02 &                0.12 &                   -0.02 &              -0.07 \\
\#9  &              0.20 &               0.08 &                0.15 &                    0.08 &               0.11 \\
\#10 &              0.19 &               0.04 &                0.15 &                    0.05 &              -0.01 \\
\#11 &              0.21 &               0.04 &                0.15 &                    0.07 &              -0.03 \\
\#12 &             -0.14 &              -0.20 &               -0.12 &                   -0.23 &              -0.25 \\
\#13 &              0.17 &              -0.01 &                0.13 &                    0.01 &              -0.06 \\
\#14 &             -0.17 &              -0.22 &               -0.16 &                   -0.23 &              -0.26 \\
\#15 &             -0.19 &              -0.23 &               -0.24 &                   -0.24 &              -0.23 \\
\#16 &             -0.19 &              -0.23 &               -0.24 &                   -0.25 &              -0.24 \\
\bottomrule\end{tabularx}
\caption{Performance of ArgumentAnalyst for selected exegetic metrics (\textbf{\scriptsize EXE-RSS}) on specific subsets (columns) of the OOD eval data.}
\label{table_main_subsets3}
\end{small}
\end{table}

\begin{table}
\begin{small}
\begin{tabularx}{\linewidth}{l *{5}{Y}}
\toprule
{} & \multicolumn{2}{c}{\emph{inference}}  & \multicolumn{2}{c}{\emph{presentation}} \\ \cmidrule(l){2-3} \cmidrule(l){4-5}
chain &  simple &  complex &  plain &  mutilat. & C\&M  \\
\midrule
\multicolumn{6}{c}{\textbf{\scriptsize EXE-JSS}} \\
\#1  &              0.06 &              -0.32 &                0.10 &                   -0.37 &              -0.37 \\
\#2  &              0.16 &              -0.17 &                0.19 &                   -0.12 &              -0.26 \\
\#3  &              0.02 &              -0.32 &                0.03 &                   -0.42 &              -0.33 \\
\#4  &              0.12 &              -0.17 &                0.13 &                   -0.14 &              -0.19 \\
\#5  &              0.15 &              -0.11 &                0.15 &                   -0.08 &              -0.18 \\
\#6  &              0.16 &              -0.14 &                0.15 &                   -0.22 &              -0.22 \\
\#7  &              0.16 &              -0.11 &                0.16 &                   -0.10 &              -0.18 \\
\#8  &              0.15 &              -0.18 &                0.14 &                   -0.19 &              -0.27 \\
\#9  &              0.23 &              -0.06 &                0.21 &                   -0.03 &              -0.21 \\
\#10  &              0.23 &              -0.12 &                0.21 &                   -0.15 &              -0.27 \\
\#11 &              0.25 &              -0.13 &                0.20 &                   -0.11 &              -0.27 \\
\#12 &              0.06 &              -0.36 &                0.04 &                   -0.28 &              -0.47 \\
\#13 &              0.13 &              -0.26 &                0.07 &                   -0.26 &              -0.40 \\
\#14 &             -0.02 &              -0.39 &               -0.07 &                   -0.31 &              -0.48 \\
\#15 &             -0.08 &              -0.41 &               -0.16 &                   -0.36 &              -0.49 \\
\#16 &             -0.08 &              -0.37 &               -0.15 &                   -0.35 &              -0.45 \\
\bottomrule\end{tabularx}
\caption{Performance of ArgumentAnalyst for selected exegetic metric (\textbf{\scriptsize EXE-JSS}) on specific subsets (columns) of the OOD eval data.}
\label{table_main_subsets4}
\end{small}
\end{table}

\begin{table}
\begin{small}
\begin{tabularx}{\linewidth}{l *{5}{Y}}
\toprule
{} & \multicolumn{2}{c}{\emph{inference}}  & \multicolumn{2}{c}{\emph{presentation}} \\ \cmidrule(l){2-3} \cmidrule(l){4-5}
chain &  simple &  complex &  plain &  mutilat. & C\&M  \\
\midrule
\multicolumn{6}{c}{\textbf{\scriptsize SYS-PP $\&$ SYS-RP $\&$ SYS-RC $\&$ SYS-US}} \\
\#1  &              0.98 &               0.78 &                1.00 &                    0.75 &               0.76 \\
\#2  &              0.97 &               0.77 &                0.99 &                    0.70 &               0.73 \\
\#3  &              0.95 &               0.79 &                0.96 &                    0.77 &               0.74 \\
\#4  &              0.95 &               0.76 &                0.96 &                    0.69 &               0.73 \\
\#5  &              0.97 &               0.75 &                0.98 &                    0.66 &               0.74 \\
\#6  &              0.96 &               0.77 &                0.98 &                    0.73 &               0.78 \\
\#7  &              0.96 &               0.73 &                0.96 &                    0.71 &               0.72 \\
\#8  &              0.97 &               0.75 &                0.97 &                    0.73 &               0.74 \\
\#9  &              0.98 &               0.80 &                0.99 &                    0.80 &               0.70 \\
\#10  &              0.98 &               0.78 &                0.99 &                    0.80 &               0.73 \\
\#11 &              0.98 &               0.78 &                0.99 &                    0.80 &               0.71 \\
\#12 &              0.97 &               0.71 &                0.97 &                    0.70 &               0.67 \\
\#13 &              0.98 &               0.81 &                0.99 &                    0.76 &               0.78 \\
\#14 &              0.96 &               0.73 &                0.96 &                    0.70 &               0.69 \\
\#15 &              0.97 &               0.72 &                0.96 &                    0.70 &               0.68 \\
\#16 &              0.97 &               0.72 &                0.96 &                    0.68 &               0.68 \\
\bottomrule
\end{tabularx}
\caption{Performance of ArgumentAnalyst for selected systematic metric (\textbf{\scriptsize SYS-PP $\&$ SYS-RP $\&$ SYS-RC $\&$ SYS-US}) on specific subsets (columns) of the OOS eval data.}
\label{table_main_subsets_oos1}
\end{small}
\end{table}

\begin{table}
\begin{small}
\begin{tabularx}{\linewidth}{l *{5}{Y}}
\toprule
{} & \multicolumn{2}{c}{\emph{inference}}  & \multicolumn{2}{c}{\emph{presentation}} \\ \cmidrule(l){2-3} \cmidrule(l){4-5}
chain &  simple &  complex &  plain &  mutilat. & C\&M  \\
\midrule
\multicolumn{6}{c}{\textbf{\scriptsize SYS-VAL}} \\
\#1  &              0.97 &               0.68 &                0.96 &                    0.74 &               0.74 \\
\#2  &              0.97 &               0.68 &                0.97 &                    0.73 &               0.71 \\
\#3  &              0.94 &               0.70 &                0.94 &                    0.72 &               0.71 \\
\#4  &              0.95 &               0.65 &                0.94 &                    0.68 &               0.71 \\
\#5  &              0.96 &               0.59 &                0.95 &                    0.65 &               0.62 \\
\#6  &              0.95 &               0.62 &                0.96 &                    0.69 &               0.63 \\
\#7  &              0.94 &               0.66 &                0.94 &                    0.66 &               0.71 \\
\#8  &              0.95 &               0.67 &                0.95 &                    0.69 &               0.69 \\
\#9  &              0.97 &               0.65 &                0.97 &                    0.72 &               0.69 \\
\#10  &              0.97 &               0.67 &                0.97 &                    0.68 &               0.72 \\
\#11 &              0.97 &               0.70 &                0.97 &                    0.68 &               0.74 \\
\#12 &              0.95 &               0.63 &                0.95 &                    0.72 &               0.70 \\
\#13 &              0.97 &               0.68 &                0.95 &                    0.73 &               0.73 \\
\#14 &              0.95 &               0.63 &                0.94 &                    0.72 &               0.69 \\
\#15 &              0.95 &               0.65 &                0.94 &                    0.75 &               0.71 \\
\#16 &              0.95 &               0.65 &                0.95 &                    0.73 &               0.71 \\
\bottomrule
\end{tabularx}
\caption{Performance of ArgumentAnalyst for selected systematic metric (\textbf{\scriptsize SYS-VAL}) on specific subsets (columns) of the OOS eval data.}
\label{table_main_subsets_oos2}
\end{small}
\end{table}

\begin{table}
\begin{small}
\begin{tabularx}{\linewidth}{l *{5}{Y}}
\toprule
{} & \multicolumn{2}{c}{\emph{inference}}  & \multicolumn{2}{c}{\emph{presentation}} \\ \cmidrule(l){2-3} \cmidrule(l){4-5}
chain &  simple &  complex &  plain &  mutilat. & C\&M  \\
\midrule
\multicolumn{6}{c}{\textbf{\scriptsize EXE-RSS}} \\
\#1  &              0.19 &              -0.16 &                0.11 &                   -0.07 &              -0.18 \\
\#2  &              0.21 &              -0.13 &                0.10 &                   -0.05 &              -0.15 \\
\#3  &              0.30 &               0.11 &                0.17 &                    0.22 &               0.06 \\
\#4  &              0.29 &               0.16 &                0.16 &                    0.24 &               0.16 \\
\#5  &              0.32 &               0.18 &                0.19 &                    0.23 &               0.18 \\
\#6  &              0.31 &               0.11 &                0.18 &                    0.19 &               0.07 \\
\#7  &              0.30 &               0.15 &                0.17 &                    0.25 &               0.16 \\
\#8  &              0.30 &               0.12 &                0.17 &                    0.24 &               0.08 \\
\#9  &              0.33 &               0.23 &                0.19 &                    0.30 &               0.23 \\
\#10  &              0.33 &               0.20 &                0.19 &                    0.27 &               0.16 \\
\#11 &              0.33 &               0.21 &                0.19 &                    0.28 &               0.16 \\
\#12 &              0.20 &               0.06 &                0.11 &                    0.16 &               0.04 \\
\#13 &              0.33 &               0.12 &                0.19 &                    0.26 &               0.07 \\
\#14 &              0.20 &               0.06 &                0.10 &                    0.16 &               0.03 \\
\#15 &              0.18 &               0.04 &                0.07 &                    0.14 &               0.00 \\
\#16 &              0.18 &               0.04 &                0.07 &                    0.11 &               0.02 \\
\bottomrule
\end{tabularx}
\caption{Performance of ArgumentAnalyst for selected exegetic metrics (\textbf{\scriptsize EXE-RSS}) on specific subsets (columns) of the OOS eval data.}
\label{table_main_subsets_oos3}
\end{small}
\end{table}

\begin{table}
\begin{small}
\begin{tabularx}{\linewidth}{l *{5}{Y}}
\toprule
{} & \multicolumn{2}{c}{\emph{inference}}  & \multicolumn{2}{c}{\emph{presentation}} \\ \cmidrule(l){2-3} \cmidrule(l){4-5}
chain &  simple &  complex &  plain &  mutilat. & C\&M  \\
\midrule
\multicolumn{6}{c}{\textbf{\scriptsize EXE-JSS}} \\
\#1  &              0.35 &              -0.14 &                0.36 &                   -0.09 &              -0.13 \\
\#2  &              0.40 &               0.02 &                0.39 &                    0.10 &               0.02 \\
\#3  &              0.30 &              -0.15 &                0.29 &                   -0.08 &              -0.15 \\
\#4  &              0.36 &               0.03 &                0.33 &                    0.08 &              -0.02 \\
\#5  &              0.41 &               0.15 &                0.39 &                    0.17 &               0.11 \\
\#6  &              0.40 &               0.04 &                0.38 &                    0.10 &              -0.01 \\
\#7  &              0.39 &               0.12 &                0.37 &                    0.15 &               0.06 \\
\#8  &              0.39 &               0.08 &                0.38 &                    0.10 &              -0.02 \\
\#9  &              0.47 &               0.16 &                0.42 &                    0.31 &               0.13 \\
\#10  &              0.47 &               0.11 &                0.42 &                    0.26 &               0.02 \\
\#11 &              0.47 &               0.11 &                0.42 &                    0.26 &               0.02 \\
\#12 &              0.40 &              -0.01 &                0.35 &                    0.14 &              -0.08 \\
\#13 &              0.45 &               0.03 &                0.36 &                    0.21 &              -0.01 \\
\#14 &              0.38 &              -0.00 &                0.30 &                    0.15 &              -0.05 \\
\#15 &              0.30 &              -0.04 &                0.22 &                    0.07 &              -0.07 \\
\#16 &              0.30 &              -0.03 &                0.22 &                    0.11 &              -0.06 \\
\bottomrule
\end{tabularx}
\caption{Performance of ArgumentAnalyst for selected exegetic metric (\textbf{\scriptsize EXE-JSS}) on specific subsets (columns) of the OOS eval data.}
\label{table_main_subsets_oos4}
\end{small}
\end{table}